\documentclass[conference]{IEEEtran}\usepackage{amsmath,amsfonts}
\usepackage{cite}
\usepackage{bookmark}
\usepackage{array}
\usepackage[caption=false,font=normalsize,labelfont=sf,textfont=sf]{subfig}
\usepackage{textcomp}
\usepackage{stfloats}
\usepackage{url}
\usepackage{verbatim}
\usepackage{graphicx}
\usepackage{amsmath}
\usepackage{multirow}
\usepackage{booktabs}
\usepackage{makecell}
\usepackage{color,soul}
\usepackage{xcolor}
\usepackage{tcolorbox}
\usepackage{setspace}
\usepackage{cite}
\usepackage{color}
\usepackage{booktabs}
\usepackage{array}
\usepackage{multirow}
\usepackage{graphicx}

\usepackage{orcidlink}
\hypersetup{
  hidelinks,
  colorlinks=true,
  allcolors=black,
  pdfstartview=Fit,
  breaklinks=true
}

\usepackage{xcolor}
\usepackage{float}
\usepackage{titlesec}
\usepackage{setspace}
\usepackage{tcolorbox}
\usepackage[framemethod=TikZ]{mdframed}
\definecolor{grayblue}{rgb}{0.4, 0.4, 0.6}

\def\BibTeX{{\rm B\kern-.05em{\sc i\kern-.025em b}\kern-.08em
    T\kern-.1667em\lower.7ex\hbox{E}\kern-.125emX}}
    
\begin{document}

\title{PanelTR: Zero-Shot Table Reasoning Framework Through Multi-Agent Scientific Discussion
}

\author{\IEEEauthorblockN{Yiran Rex Ma \orcidlink{0009-0002-3440-9819}}
\IEEEauthorblockA{\textit{School of Humanities}\\ 
\textit{Beijing Univeristy of Posts and Telecommunications}\\
Beijing, China \\
mayiran@bupt.edu.cn}
}

\maketitle

\begin{abstract}
Table reasoning, including tabular QA and fact verification, often depends on annotated data or complex data augmentation, limiting flexibility and generalization. LLMs, despite their versatility, often underperform compared to simple supervised models. To approach these issues, we introduce PanelTR, a framework utilizing LLM agent scientists for robust table reasoning through a structured scientific approach. PanelTR's workflow involves agent scientists conducting individual investigations, engaging in self-review, and participating in collaborative peer-review discussions. This process, driven by five scientist personas, enables semantic-level transfer without relying on data augmentation or parametric optimization. Experiments across four benchmarks show that PanelTR outperforms vanilla LLMs and rivals fully supervised models, all while remaining independent of training data. Our findings indicate that structured scientific methodology can effectively handle complex tasks beyond table reasoning with flexible semantic understanding in a zero-shot context. \footnote{Code, prompts and evaluation scripts are openly available at \\ \url{https://github.com/rexera/PanelTR}}
\end{abstract}

   
\section{Introduction}
Table reasoning, including fact verification\cite{aly2021feverous, wang2021semeval} and question answering\cite{zhu2021tatqa}, is crucial for automated information processing across various domains \cite{herzig2020tapas, liu2021tapex}. It is essential for extracting, validating, and synthesizing structured information from scientific literature, financial reports, medical records, and technical documentation \cite{zhu2021tat, chen2020tabfact}. As organizations handle increasing volumes of structured data, accurate table reasoning becomes more critical.

Table reasoning has evolved from rule-based systems to deep learning methods \cite{pasupat2015compositional, zhong2017seq2sql}. Supervised learning shows strong performance but demands extensive annotated data, which is costly for new domains \cite{herzig2020tapas, chen2020hybridqa}. Unsupervised methods offer flexibility through data augmentation and self-training \cite{varshney2021unsupervised, dong2023toward}, yet require significant engineering and expertise. Both approaches have advanced the field: supervised methods excel in specific tasks, while unsupervised methods adapt to new scenarios with less annotation \cite{li2023toward, chemmengath2021topic}. However, challenges in semantic understanding hinders their effectiveness in novel or complex scenarios \cite{eisenschlos2020understanding}. These issues drive research towards more robust, adaptable solutions that balance performance and practicality.

Despite the advancements in large language models (LLMs) and their capabilities in natural language processing\cite{openai2024gpt4ocard, deepseekai2024deepseekv3technicalreport}, challenges in table reasoning remain. LLMs excel in Chain-of-Thought (CoT) prompting\cite{wei2022chain}, few-shot learning\cite{brown2020language}, and zero-shot reasoning\cite{kojima2022large}, but often underperform in structured table reasoning compared to supervised methods. This is due to their tendency for instant responses without systematic investigation, inconsistencies in numerical reasoning, and difficulties with complex multi-step operations in table analysis.

To approach to these limitations, we propose PanelTR, an LLM-backed multi-agent system (MAS) that leverages agent scientists for robust table reasoning through structured scientific methodology - self-review and peer-review. Our approach focuses on developing a systematic, plug-and-play workflow that complements existing LLM rather than advancing neural network architectures.  The framework operates in three key phases: individual \textit{Investigation}, \textit{Self-Review}, and \textit{Peer-Review}, mimicking the rigorous process of scientific inquiry. Using five specialized LLM scientist personas, we demonstrate how existing LLM capabilities can be effectively harnessed for complex table reasoning. Through orchestrating this analytical process on four distinct table reasoning tasks, PanelTR achieves superior reasoning capabilities through flexible semantic understanding without requiring task-specific training data or complex augmentation strategies, achieving zero-shot transfer. Our main contributions include:

\begin{itemize}
\item We introduce PanelTR, a framework for table reasoning using LLM agent scientists. This method enables robust reasoning through systematic investigation, self-review, and peer-review, without relying on training data or task-specific adaptations.

\item We show that existing LLM capabilities can be enhanced through scientific methodology without changing their architecture. The multi-agent design allows for comprehensive analysis and complex reasoning.

\item Our experiments across multiple benchmarks demonstrate that PanelTR achieves competitive performance while remaining independent of task-specific training data, highlighting its zero-shot transferability.
\end{itemize}

\section{Related Works}

\subsection{Tabular Reasoning Models}

\textit{Symbolic Execution and SQL-based Reasoning} characterized early table reasoning through semantic parsing that converted natural language to executable programs \cite{guu2017language, liang2016neural}. These approaches worked well for structured data like relational databases, with Seq2SQL \cite{zhong2017seq2sql} successfully translating natural language to SQL queries. However, these methods were highly task-specific and struggled to generalize beyond narrow domains. While models like TabSQLify \cite{nahid2024tabsqlify} and H-STAR \cite{abhyankar2024hstar} attempted improvements through query decomposition, they remained limited by strict programmatic reasoning that produced brittle, task-specific solutions\cite{yang2020program}.

\textit{Neural Architectures and Graph-Based Approaches} incorporate table structure into learned representations through graph-based methods. RegHNT \cite{lei2022reghnt} encodes relationships between cells and text using dynamic graph reasoning for complex hybrid datasets. TAG-QA \cite{zhao2023tagqa} extends this through graph-ranking for answer generation, while TaCube \cite{zhou2022tacube} optimizes performance using pre-computed operations.

\textit{Pre-Trained Language Models (PLMs)} have enhanced table reasoning by leveraging auxiliary knowledge and fine-tuned representations \cite{glass2021capturing, duan2021bridging, duan2022not, li2021jointly}. Notable examples include TAPAS \cite{herzig2020tapas}, which extended BERT for table encoding, and GraPPa \cite{yu2020grappa}, which optimized learning through text-schema linking. However, PLM-based approaches require large annotated, in-domain datasets for pre-training, limiting transfer potentials.

\textit{Unsupervised Data Generation and Self-Training} approaches have emerged for table reasoning without labeled datasets \cite{varshney2021unsupervised, dong2023toward, li2023toward}. SQL-based synthesis \cite{chemmengath2021topic} and context-free grammars \cite{eisenschlos2020understanding} generate synthetic data but struggled with generalization. Frameworks like FlexKBQA and UCTR \cite{li2024flexkbqa, uctr-st} combine synthetic data with few-shot learning and self-training \cite{li2022effective}, with dependence on initial warming-up limiting effectiveness because of cold start issue.

\textit{CoT Reasoning with LLMs} has introduced new approaches by integrating CoT prompting into table reasoning. LLM-based models generate natural language reasoning steps for flexible analysis. Chain-of-Table \cite{wang2024chainoftable} tracks tabular transformations through iterative evolution, enabling dynamic operation planning. FLEXTAF \cite{zhang2024flextaf} enhances this by selecting optimal representations based on question complexity, choosing between structured retrieval and logical deduction. ReAct \cite{yao2022react} further advances by interleaving reasoning and action steps, allowing models to dynamically plan and execute reasoning processes. 

\textit{Multi-Agent and Hybrid methods} have emerged to decompose table reasoning into specialized subtasks. MACT \cite{zhou2024mact} uses collaborating agents to break down and execute reasoning steps while tracking the process explicitly. TabLaP \cite{wang2024tablpa} separates numerical computation from language understanding to improve accuracy on arithmetic-heavy tasks. DEPS \cite{wang2023deps} introduces a novel framework for dynamic expert selection and collaboration, showing how specialized agents can complement each other's strengths. Reflexion \cite{shinn2023reflexion} enhances this by incorporating self-reflection mechanisms that allow agents to learn from their mistakes and improve their reasoning strategies over time. 

\subsection{Datasets for Table Reasoning}

\textit{Table-Only Datasets} focus on structured tabular data, requiring models to understand and reason over tables. A prominent example is WikiTableQuestions (WikiTQ) \cite{pasupat2015compositional}, which contains natural language questions about Wikipedia tables and requires free-form text generation. TabFact \cite{chen2020tabfact} tests fact verification by determining if claims are supported by tables, evaluating logical reasoning capabilities including negation and numerical comparisons \cite{zhong2020logicalfactchecker}.

\textit{Hybrid Table-Text Datasets} combine structured and unstructured data sources. HybridQA \cite{chen2020hybridqa} tests multi-hop reasoning by linking tables with text descriptions, requiring models to extract entities and cross-reference information. InfoboxQA \cite{kumar2022infoboxqa} extends this through knowledge graph-style entity linking.

\textit{Domain-Specific Datasets} evaluate reasoning in specialized fields. TAT-QA \cite{zhu2021tatqa} tests financial reasoning with arithmetic operations \cite{li2023financialqa}, while MultiHiertt \cite{zhao2022multi} focuses on hierarchical financial tables. SciTabQA \cite{ghosh2024scitabqa} evaluates scientific table reasoning \cite{verga2020scierc}, while recent benchmarks like TableBench \cite{wu2024tablebench} incorporate diverse tabular data to evaluate complex multi-step and multi-modal reasoning.

\section{PanelTR}
We propose PanelTR, a scientific analysis framework that approaches table reasoning through structured scientist investigation and collaborative discussion. Drawing inspiration from scientific methodology and panel review processes, our framework implements three core stages: individual \textit{Investigation}, methodical \textit{Self-Review}, and collective scientist \textit{Peer-Review}.

\subsection{Problem Definition}

Table reasoning encompasses tasks that require analyzing and reasoning over tabular data. Given a table $\tau$ and its surrounding context $P$ (any additional textual information), along with a natural language query $\xi$ , the goal is to generate an appropriate response $y$  based on task requirements. The output $y$ varies by task - it could be a natural language answer, verification label, SQL transcription, or other task-specific responses. Formally, for table-only scenarios, we aim to construct a mapping function $f$ such that $y = f(\tau, \xi)$. For cases involving both tabular and textual evidence, the mapping extends to $y = f(\tau, P, \xi)$, where $y$ must be logically consistent with both evidence and task requirements.

The key challenge lies in achieving robust reasoning that can handle diverse table structures while maintaining strong reasoning capabilities across tasks. The system must effectively parse tabular structures ranging from simple relational tables to complex hierarchical formats, integrate information from both structured tables and unstructured text when necessary, and execute appropriate reasoning steps based on specific task requirements.

\begin{figure*}[t]
\centering
\includegraphics[width=0.9\textwidth,page=1]{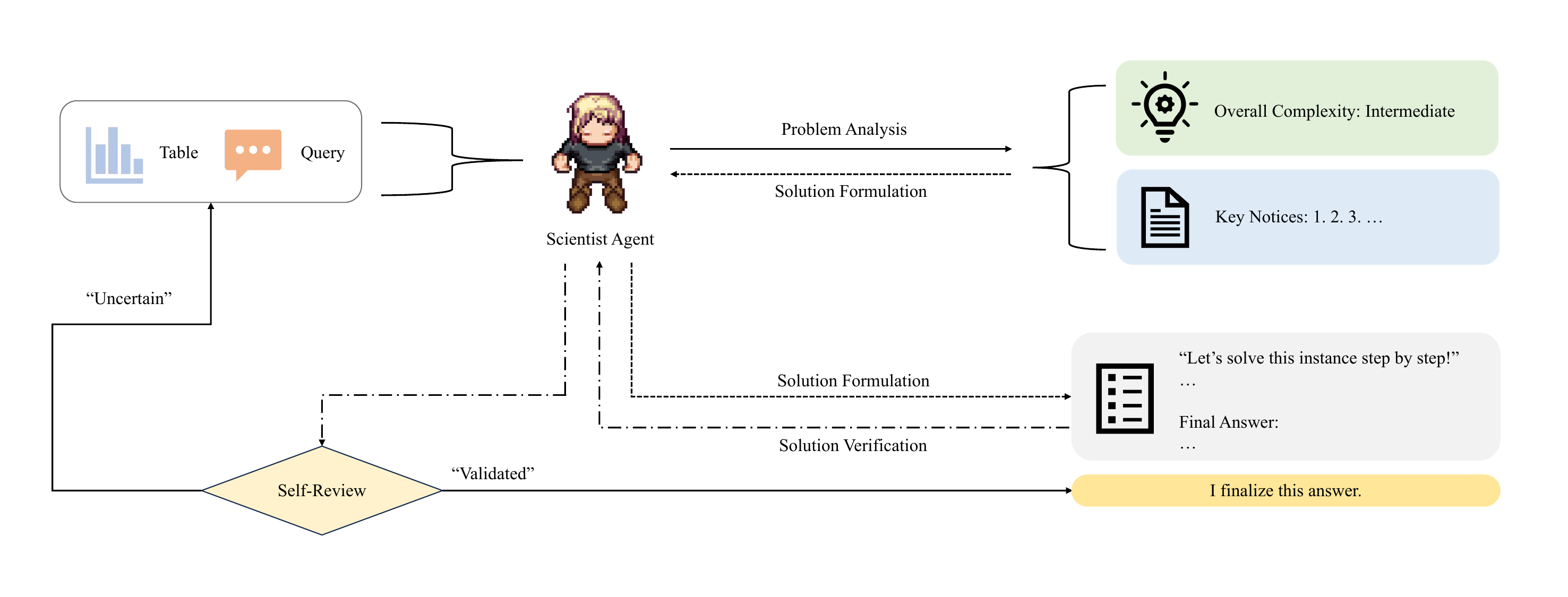}
\caption{The \textit{Investigation} and \textit{Self-Review} stages of individual scientist agents. The figure illustrates the systematic workflow where each scientist agent: (I) \textit{Problem Analysis} - assesses problem complexity and identifies key insights; (II) \textit{Solution Formulation} - proposes an initial solution strategy based on the analysis; (III) \textit{Self-Review} - verifies the solution's validity.}
\label{fig:individual}
\end{figure*}

\subsection{Investigation}

\textit{Investigation} includes (as shown in Fig. \ref{fig:individual}): (I) \textit{Problem Analysis} - assessing task difficulty and identifying critical reasoning notices; (II) \textit{Solution Formulation} - developing a comprehensive strategy. Each scientist agent $\phi$ initially evaluates the query $\xi$ and table $\tau$ through an assessment function $\mathcal{A}: \mathcal{X} \rightarrow \mathcal{D} \times \mathcal{G}$, where $\mathcal{X}$ represents the input space containing all possible table-query pairs, $\mathcal{D}$ represents the complexity space, and $\mathcal{G}$ represents the analytical points space:

\[
\delta, \gamma = \mathcal{A}(\xi, \tau)
\]

where $\delta \in \mathcal{D} = \{\text{basic}, \text{intermediate}, \text{complex}\}$ represents the assessed problem complexity and $\gamma \in \mathcal{G}$ identifies key analytical points (highlighting specific aspects of the problem that require attention). Based on this assessment, the scientist formulates a preliminary solution strategy through a strategy function $\mathcal{S}: \mathcal{D} \times \mathcal{G} \rightarrow \mathcal{Y}$, where $\mathcal{Y}$ represents the solution space containing all possible answers:

\[
\sigma_{\text{init}} = \mathcal{S}(\delta, \gamma)
\]

where $\sigma_{\text{init}}$ represents the initial solution proposed by the scientist. The investigation adapts to problem complexity - for numerical comparison tasks, $\delta$ might indicate ``intermediate'' complexity with $\gamma$ highlighting requirements like unit standardization, while simpler retrieval queries would be classified as ``basic'' with $\gamma$ focusing on intuitive data location. After formulating an initial solution, the scientist proceeds to rigorously validate their findings through self-review.

\subsection{Self-Review}

At each iteration, the scientist evaluates the current solution through a verification function $\mathcal{V}: \mathcal{Y} \times \mathcal{X} \rightarrow \Omega$, where $\Omega$ represents the confidence space:

\[
\omega = \mathcal{V}(\sigma, \xi, \tau)
\]

where $\omega \in \Omega = \{\text{``uncertain''}, \text{``validated''}\}$ indicates confidence level (representing the scientist's assessment of the solution's reliability). A solution is marked ``uncertain'' when methodological gaps or inconsistencies exist, achieving ``validated'' status only when demonstrating robust consistency with both query requirements and evidence.

When ``uncertain'', the self-review process reuses the assessment function $\mathcal{A}$ and solution function $\mathcal{S}$ in an iterative manner. When uncertainty remains, the scientist continues to refine the solution by iteratively applying the assessment function $\mathcal{A}$, solution function $\mathcal{S}$, and verification function $\mathcal{V}$:

\[
\sigma^{(t)} = \mathcal{S}(\mathcal{A}(\xi, \tau)), \quad \omega^{(t)} = \mathcal{V}(\sigma^{(t)}, \xi, \tau), \quad t \leq t_{\text{max}}
\]

where $\sigma^{(t)}$ represents the refined solution at iteration $t$, and $t_{\text{max}}$ denotes the maximum number of refinement iterations allowed. This validation process continues until either achieving ``validated'' or reaching maximum iterations $t_{\text{max}}$. After completing individual \textit{Investigation} and \textit{Self-Review}, scientists proceed to engage in collaborative \textit{Peer-Review} discussions.

\begin{figure*}[t]
\centering
\includegraphics[width=0.9\textwidth,page=2]{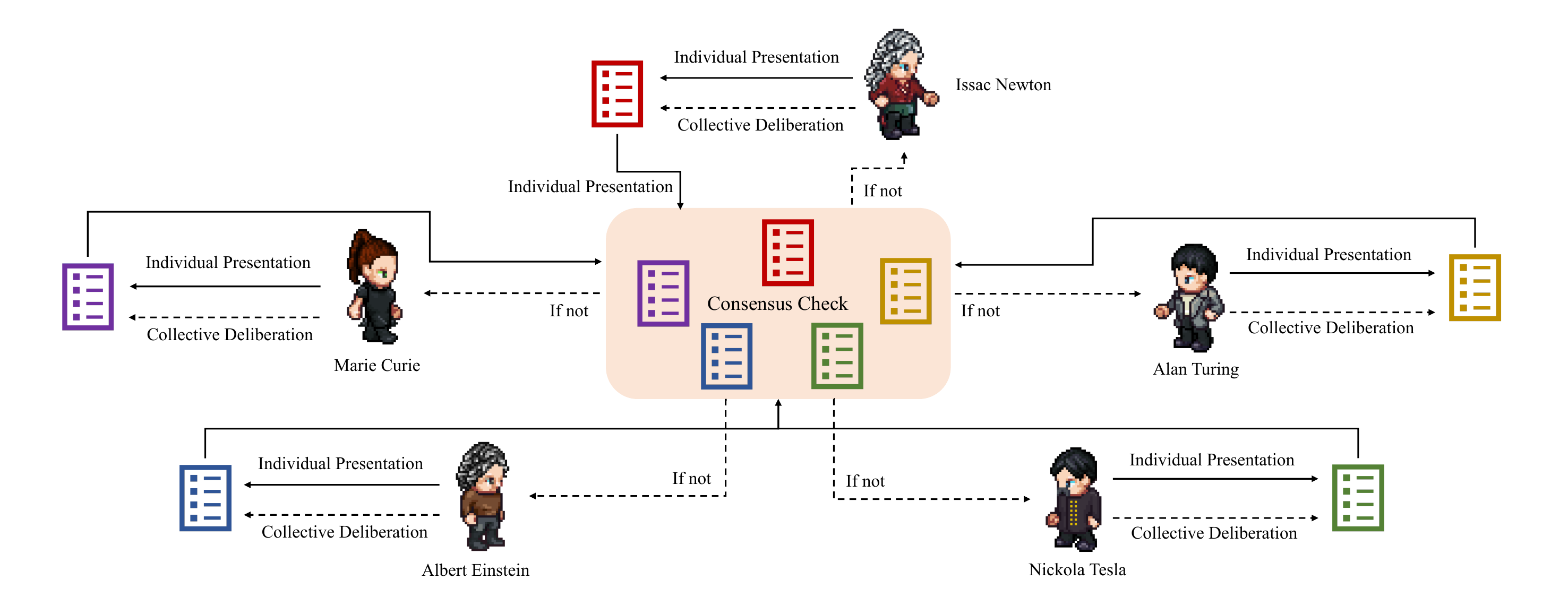}
\caption{The \textit{Peer-Review} process in multi-agent panel discussions. Each scientist performs \textit{Individual Presentation}, followed by iterative refinement from \textit{Collective Deliberation} if no consensus. The process iterates until either consensus or maximum iteration is reached, at which point majority voting is used.}
\label{fig:panel}
\end{figure*}

\subsection{Peer-Review}

\textit{Peer-Review} (as shown in Fig. \ref{fig:panel}) facilitates a collaborative scientific review through structured discussions among five distinguished scientists, each contributing unique analytical perspectives (prompt designs are detailed in Table \ref{tab:personas}): \textit{Albert Einstein} (E); \textit{Isaac Newton} (N); \textit{Marie Curie} (M); \textit{Alan Turing} (A); and \textit{Nikola Tesla} (T). Let $\Phi = \{E, N, M, A, T\}$ represent the set of scientist agents, each with a unique expertise function $\phi_r: \mathcal{X} \rightarrow \mathcal{Y}$ where $r \in \Phi$ (reusing \textit{Investigation} and \textit{Self-Review} for individual scientists). Each scientist independently analyzes and presents their solutions to the panel in a random order (\textit{Individual Presentation}). Scientists presenting later can observe previous presentations and adjust their solutions accordingly:

\[
\sigma_r^{\text{init}} = \phi_r(\xi, \tau), \quad r \in \Phi
\]

where $\sigma_r^{\text{init}}$ denotes the initial solution proposed by scientist $r$. If all solutions are identical, indicating unanimous agreement, the final solution is:

\[
\sigma_{\text{final}} = \sigma_r^{\text{init}} \quad \text{if} \quad \forall r,s \in \Phi, \sigma_r^{\text{init}} = \sigma_s^{\text{init}}
\]

Otherwise, the panel engages in iterative structured discussion rounds, where scientists can either modify their solution based on peer feedback or maintain their original position before presenting (\textit{Collective Deliberation}). At each iteration $t$, each scientist presents their solution $\sigma_r^{(t)}$. The process continues until either consensus is reached or the maximum number of iterations $t_{\text{max}}$ is exhausted:

\[
\sigma_{\text{final}} = \sigma_r^{(t)} \quad \text{if} \quad \forall r,s \in \Phi, \sigma_r^{(t)} = \sigma_s^{(t)} \quad \text{and} \quad t \leq t_{\text{max}}
\]

If consensus remains elusive after $t_{\text{max}}$ iterations, the final solution is determined through majority voting:

\[
\sigma_{\text{final}} = \text{MajorityVoting}(\{\sigma_r^{(t_{\text{max}})} \mid r \in \Phi\})
\]

Through this structured process of \textit{Individual Presentation} and \textit{Collective Deliberation}, the panel effectively combines diverse scientific perspectives to reach well-examined solutions, completing the three-stage PanelTR framework.

\begin{table}[htbp]
\centering
\caption{Scientist Personas and Their Prompt Design}\scalebox{0.65}{
\begin{tabular}{ll}
\toprule
\textbf{Scientist} & \textbf{Prompt Focus} (You should...)\\
\midrule
Albert Einstein & Explore alternative interpretations and conceptual frameworks \\
\midrule
Isaac Newton & Verify numerical relationships and logical consistency \\
\midrule
Marie Curie & Validate with experimental evidence and practical tests \\
\midrule
Alan Turing & Analyze problem structure and optimize solution efficiency \\
\midrule
Nikola Tesla & Synthesize diverse perspectives into coherent solutions \\
\bottomrule
\end{tabular}}
\label{tab:personas}
\end{table}

\section{Experiments}

\subsection{Datasets}
To evaluate the effectiveness of PanelTR, we conduct experiments across four representative benchmarks. FEVEROUS \cite{aly2021feverous} focuses on fact verification (deciding on whether system \textit{Supports} or \textit{Refutes} the statement, or responds with \textit{Not Enough Information (NEI)})  using evidence derived from Wikipedia, comprising both textual and tabular data. TAT-QA \cite{zhu2021tat} targets question answering with hybrid evidence sourced from real-world financial reports. WiKiSQL \cite{zhong2017seq2sql} consists of question-SQL query pairs over Wikipedia tables, aiming at natural language query transcription. Finally, SEM-TAB-FACTS \cite{wang2021semeval} addresses fact verification (classifying claims as \textit{Entailed}, \textit{Refuted}, or \textit{Unknown}) using tables extracted from scientific articles. 

\subsection{Experiment Setup}

We utilized \texttt{deepseek-v3}\cite{deepseekai2024deepseekv3technicalreport} as our foundational model\footnote{https://api.deepseek.com, ``deepseek-chat''} with default temperature (1.0) to evaluate the overall performance of PanelTR across the four benchmarks. For data preprocessing, we flattened all tabular inputs into string format to better accommodate the LLM's natural language capabilities. We curated task description of each benchmark for vanilla LLMs and our scientists. For \textit{Investigation} and \textit{Peer-Review}, we set the maximum number of iterations to 1 ($t_{\text{max}}=1$). While more iterations could theoretically lead to better convergence, our empirical results showed that a single iteration achieved optimal performance by avoiding oscillating responses that could degrade accuracy.

Our prompts incorporated several key principles: (1) assigning exact one atomic prompt to function $\mathcal{A,S,V}$, \textit{Individual Presentation}, and \textit{Collective Deliberation}, which are further implemented by nested response generation calls for process memory, (2) encouraging scientists to maintain healthy skepticism about problem difficulty rather than defaulting to oversimplified assessments, and (3) promoting flexible and diverse perspectives by allowing panel members to freely modify personal solutions or maintain dissenting views rather than forcing artificial consensus. These prompt engineering choices helped create more nuanced and robust multi-agent discussions. All intermediate results are regex-extracted by fixed format requirements in prompts, making the entire workflow hard-wired using \texttt{AutoGen v0.2.40} \cite{wu2024autogen}.

\subsection{Baselines}

We compare PanelTR against several strong baselines across multiple datasets. For supervised approaches: TAGOP \cite{zhu2021tat} identifies relevant table cells and text spans before applying predefined operators; FinMath \cite{li_finmath_2022} integrates a tree-structured solver for multi-step numerical reasoning over financial reports, improving arithmetic QA; NumNet \cite{ran_numnet_2019} introduces a numerically-aware graph neural network to explicitly model numerical relations; UniPCQA \cite{deng_pacific_2023} unifies Proactive Conversational QA over financial tables and text via a Seq2Seq framework, reformulating numerical reasoning as code generation for enhanced arithmetic consistency; FEVEROUS baseline \cite{aly2021feverous} combines a retriever module for evidence extraction with a verdict predictor for final classification; TAPAS \cite{herzig2020tapas} employs specialized positional embeddings and joint pre-training on textual and tabular data, showing strong performance; TAPAS-Transfer \cite{chen2019tabfact} leverages transfer learning from TABFACT for table fact verification; and TAPEX \cite{liu2021tapex}, a generative pre-trained model mimicking neural SQL executor. For unsupervised and few-shot approaches: MQA-QG \cite{pan2020unsupervised} generates questions/claims by identifying bridge entities between tables and text; vanilla \texttt{gpt-4o-mini} \cite{openai2024gpt4ocard} and \texttt{deepseek-v3} \cite{deepseekai2024deepseekv3technicalreport} serve as foundation model baselines generating answers directly from task instructions; and UCTR-ST \cite{uctr-st} employs program generation and transformation modules to produce synthetic training data, utilizing 50 labeled instances in few-shot settings for iterative model fine-tuning.

\subsection{Metrics}
Evaluation metrics are tailored to each benchmark. For TAT-QA, we use \textit{Exact Match (EM)} and \textit{F1 score}. WiKiSQL uses \textit{denotation accuracy} to match predicted answers with the ground truth. SEM-TAB-FACTS employs a three-way \textit{micro F1 score} for classifying claims as Supported, Refuted, or Unknown. FEVEROUS involves evidence retrieval and reasoning, measured by \textit{label accuracy} and the stricter \textit{FEVEROUS score}, which requires correct evidence and classification. Our focus on reasoning uses baseline retrieval from \cite{aly2021feverous} for evaluation on the development set.

\begin{table}[htbp]
  \begin{center}
  \caption{Results on TAT-QA dev set}\scalebox{0.7}{
  \begin{tabular}{c|l|rr}
  \toprule
  \multicolumn{2}{c|}{Model} & EM & F1 \\
\midrule
\multirow{5}{*}{Supervised}
& TAPAS\cite{herzig2020tapas} & 18.9 & 26.5 \\
& NumNet+ \cite{ran_numnet_2019} & 38.1 & 48.3 \\
& TAGOP\cite{zhu2021tat} & 55.5 & 62.9 \\ 
& FinMath\cite{li_finmath_2022} & 60.5 & 66.3 \\
& UniPCQA\cite{deng_pacific_2023} & 64.7 & 72.0 \\
\midrule
\multirow{2}{*}{Few-Shot}
& TAGOP\cite{zhu2021tat} & 8.3 & 12.1 \\
& TAGOP+UCTR-ST\cite{uctr-st} & 48.1 & 56.9 \\ 
\midrule
\multirow{5}{*}{Unsupervised}
& MQA-QG \cite{pan2020unsupervised}& 19.4 & 27.7 \\
& UCTR-ST\cite{uctr-st} & 40.2 & 47.6 \\
& \texttt{gpt-4o-mini}\cite{openai2024gpt4ocard} & 37.0 & 42.8 \\
& \texttt{deepseek-v3}\cite{deepseekai2024deepseekv3technicalreport} & \underline{58.0} & \underline{66.5} \\ 
& \textbf{PanelTR (ours)} & \textbf{67.2} & \textbf{74.8} \\
\bottomrule
\end{tabular}}
\label{tatqa}
\end{center}
\end{table}

\begin{table}[htbp]
  \centering
  \caption{Results on SEM-TAB-FACTS}\scalebox{0.7}{
  \begin{tabular}{c|l|rr}
  \toprule
  \multicolumn{2}{c|}{Model}  & Dev& Test \\
\midrule
\multirow{1}{*}{Supervised}
& TAPAS\cite{herzig2020tapas} & 66.7 & 62.4 \\
\midrule
\multirow{2}{*}{Few-Shot}
& TAPAS\cite{herzig2020tapas} & 48.6 & 46.5 \\
& TAPAS+UCTR-ST\cite{uctr-st} & 64.1 & 61.0 \\
\midrule
\multirow{6}{*}{Unsupervised}
& MQA-QG \cite{pan2020unsupervised} & 53.2 & 50.4 \\
& TAPAS-Transfer\cite{chen2019tabfact} & 59.0 & 58.7 \\
& UCTR-ST\cite{uctr-st} & 64.2 & 61.2 \\
& \texttt{gpt-4o-mini}\cite{openai2024gpt4ocard} & 71.8 & 71.4 \\
& \texttt{deepseek-v3}\cite{deepseekai2024deepseekv3technicalreport} & \underline{74.3} & \underline{83.3} \\
& \textbf{PanelTR (ours)} & \textbf{87.1} & \textbf{90.8} \\
\bottomrule
\end{tabular}}
\label{semtab}
\end{table}

\begin{table}[htbp]
  \centering
  \caption{Results on WiKiSQL}\scalebox{0.7}{
  \begin{tabular}{c|l|rr}
  \toprule
  \multicolumn{2}{c|}{Model} & Dev & Test \\
\midrule
\multirow{2}{*}{Supervised}
& TAPAS\cite{herzig2020tapas} & 85.1 & 83.6 \\
& TAPEX\cite{liu2021tapex} & \textbf{88.1} & \underline{87.0} \\
\midrule
\multirow{2}{*}{Few-Shot}
& TAPEX\cite{liu2021tapex} & 53.8 & 52.9 \\
& TAPEX+UCTR-ST\cite{uctr-st} & 63.5 & 62.7 \\
\midrule
\multirow{6}{*}{Unsupervised}
& TAPEX\cite{liu2021tapex} & 21.4 & 21.8 \\
& MQA-QG \cite{pan2020unsupervised} & 57.8 & 57.2 \\
& UCTR-ST\cite{uctr-st} & 63.5 & 62.7 \\
& \texttt{gpt-4o-mini}\cite{openai2024gpt4ocard} & 79.5 & 78.5 \\
& \texttt{deepseek-v3}\cite{deepseekai2024deepseekv3technicalreport} & 85.6 & 85.4 \\
& \textbf{PanelTR (ours)} & \underline{87.2} & \textbf{87.3} \\
\bottomrule
\end{tabular}}
\label{wikisql}
\end{table}

\begin{table}[htbp]
  \centering
  \caption{Results on FEVEROUS dev set}\scalebox{0.7}{
  \begin{tabular}{c|l|rr}
  \toprule
  \multicolumn{2}{c|}{Model} & Acc & Score \\
\midrule
\multirow{2}{*}{Supervised}
& Table-only baseline\cite{aly2021feverous} & \underline{81.6} & 19.1 \\
& Full baseline\cite{aly2021feverous} & \textbf{86.0} & 20.2 \\
\midrule
\multirow{2}{*}{Few-Shot}
& Full baseline\cite{aly2021feverous} & 67.3 & 14.2 \\
& Full baseline+UCTR-ST\cite{uctr-st} & 78.2 & 19.7 \\
\midrule
\multirow{5}{*}{Unsupervised}
& MQA-QG \cite{pan2020unsupervised} & 71.1 & 17.6 \\
& UCTR-ST\cite{uctr-st} & 77.7 & 19.7 \\
& \texttt{gpt-4o-mini}\cite{openai2024gpt4ocard} & 72.5 & 23.2 \\
& \texttt{deepseek-v3}\cite{deepseekai2024deepseekv3technicalreport} & 74.6 & \underline{23.5} \\
& \textbf{PanelTR (ours)} & 75.5 & \textbf{24.1} \\
\bottomrule
\end{tabular}}
\label{feverous}
\end{table}

\subsection{Results}

\paragraph{Overall Performance.}
As shown in Tables \ref{tatqa}–\ref{feverous}, PanelTR demonstrates competitive performance across all benchmarks, surpassing both supervised and unsupervised baselines on TAT-QA. For SEM-TAB-FACTS, PanelTR shows significant improvement over existing baselines. While WikiSQL results are slightly below the fully supervised TAPEX model, likely due to the task's highly structured nature requiring precise SQL syntax, PanelTR still demonstrates strong zero-shot performance. This performance stems from flexible understanding and scientific deliberation, while traditional approaches are comparatively rigid or lacking in effective inference accumulation.

\paragraph{Trade-Offs in FEVEROUS.}
PanelTR excels in FEVEROUS scores but, like LLMs such as \texttt{gpt-4o-mini} and \texttt{deepseek-v3}, falls short of supervised baselines in label accuracy. This is due to the FEVEROUS score's emphasis on complex reasoning over simple accuracy. Supervised methods may excel in straightforward cases but struggle with complex ones, while LLMs and PanelTR handle complex cases better but may accumulate bias in simpler questions. A major challenge in FEVEROUS is the fragmented evidence from the baseline retriever, as complete tables offers benefits but exceeds model context limits. Future work could integrate structured frameworks with LLMs to improve both accuracy and reasoning.

\subsection{Ablation Study \& Discussion}

\paragraph{Effectiveness of Investigation} 
Table \ref{tab:component} demonstrates that incorporating the \textit{Investigation} stage leads to consistent improvements over the vanilla model across most benchmarks. This methodology, which resembles a more structured form of zero-shot Chain-of-Thought prompting, provides agents with a scientific approach to assess problem complexity and develop appropriate solution strategies. The most pronounced improvements appear in TAT-QA and SEM-TAB-FACTS, where complex reasoning paths benefit from explicit problem decomposition. Although FEVEROUS shows a moderate decrease in accuracy, this may indicate the task's sensitivity to overanalysis when statements require straightforward verification. The investigation stage establishes a critical foundation for the framework by enabling more methodical reasoning compared to direct inference.

\begin{table}[htbp]
  \centering
  \caption{Component Ablation Results}\scalebox{0.7}{
  \begin{tabular}{l|cccc}
  \toprule
  \textbf{Strategy} & \textbf{TAT EM} & \textbf{FEV Acc} & \textbf{SEM Dev} & \textbf{Wiki Dev} \\
  \midrule
  Vanilla & 58.0 & 74.6 & 74.3 & 85.6 \\
  Vanilla+\textit{Self} & 60.7 & 72.1 & 78.5 & 86.1 \\
  Vanilla+\textit{Peer} & 62.1 & 69.6 & 79.8 & 85.4 \\
  Vanilla+\textit{Self}+\textit{Peer} & 64.5 & 68.1 & 81.0 & 86.7 \\
  \textit{Investigation} & 61.6 & 71.3 & 78.2 & 85.8 \\
  \textit{Investigation}+\textit{Self} & 65.7 & 58.7 & 84.3 & 83.2 \\
  \textit{Investigation}+\textit{Peer} & 65.4 & 62.5 & 85.6 & 85.3 \\
  \textbf{PanelTR} (Complete) & 67.2 & 73.0 & 87.1 & 87.2 \\
  - Random Role& 66.8 & 72.2 & 87.4 & 86.8 \\
  - Alternative Role & 67.0 & 72.9 & 86.9 & 86.5 \\
  \bottomrule
  \end{tabular}}
  \label{tab:component}
\end{table}

\begin{figure}[ht]
    \centering
    \includegraphics[width=0.75\linewidth]{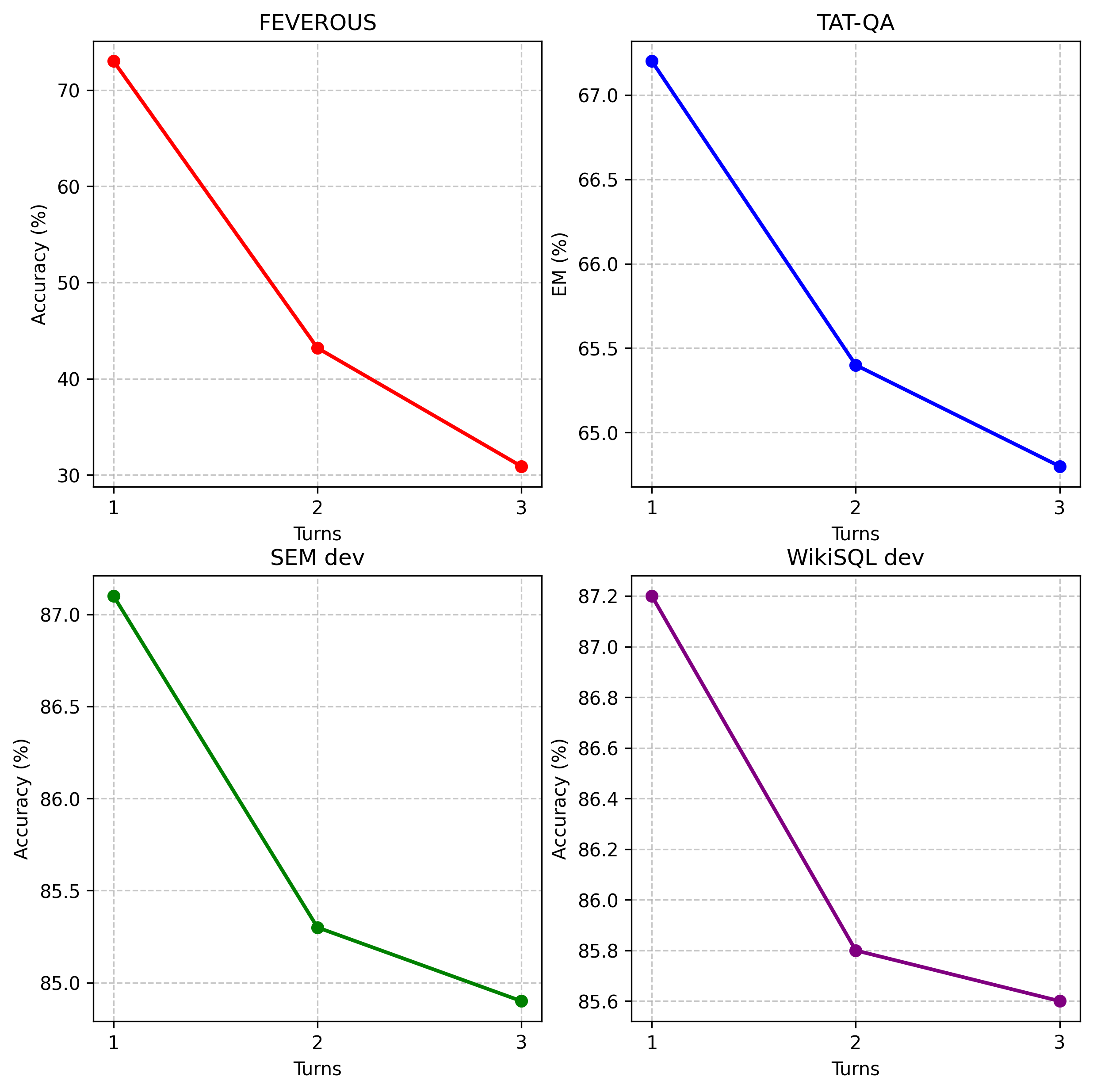}
    \caption{Iteration Turn Study on All Benchmarks.}
    \label{fig:turns}
\end{figure}

\paragraph{Impact of Self-Review}
\textit{Self-Review} demonstrates substantial positive effects across multiple configurations. When applied to the vanilla model, it produces improvements in three benchmarks, with only FEVEROUS showing a moderate decline. More significantly, when combined with the Investigation stage, it yields even more substantial gains in two. This suggests a synergistic relationship between structured problem analysis and iterative self-validation. The notable exception is FEVEROUS, where adding \textit{Self-Review} to investigation causes a considerable performance drop. This indicates that excessive analytical iteration may introduce uncertainty in verification tasks where confidence plays a crucial role. In numerical reasoning tasks like in TAT-QA, however, \textit{Self-Review} contributes significantly to accuracy by encouraging scientists to verify calculations and validate assumptions before presenting conclusions.

\paragraph{Contribution of Peer-Review}
\textit{Peer-Review} shows mixed results that vary by task type and starting configuration. When added to the vanilla model, it yields substantial improvements in TAT-QA and SEM-TAB-FACTS while causing a decline in FEVEROUS. The same applies for \textit{Investigation} stage. This pattern suggests that multi-perspective, collaborative discussion particularly benefits tasks requiring complex reasoning and interpretation, while potentially introducing confusion in straightforward classification tasks. Despite these variations, the complete PanelTR framework's superior performance indicates that \textit{Peer-Review}, when properly integrated with \textit{Investigation} and \textit{Self-Review}, contributes positively to the overall framework.

\paragraph{Influence of Scientist Roles}
To assess whether our specific scientist roles drive performance improvements, we conducted experiments with alternative role configurations. The Random Role configuration (using 2-5 randomly selected scientists from our pool) and Alternative Role setup (using five different professions: Doctor, Artist, Researcher, Social Influencer, and Entrepreneur) both achieve comparable performance to the complete PanelTR across all benchmarks. This suggests that the benefits of our framework derive from the structured scientific approach and diverse perspective integration rather than from specific persona choices. This finding emphasizes the robustness of our scientific methodology as the primary driver of performance improvements rather than specific role selection or role quantity.

\paragraph{Iteration Study: Less is More} 

Interestingly, increasing the maximum iterations in panel discussion leads to performance degradation, particularly on fact verification tasks, as shown in Fig. \ref{fig:turns}. This suggests that excessive iterations between multiple agents might dilute the benefits of collaborative reasoning, especially when dealing with clear-cut scenarios, which calls for a balance between spontaneous inference and deliberative collective wisdom.

\paragraph{Limitations}

PanelTR's heavy reliance on pre-trained LLMs limits its ability to develop novel reasoning beyond the base model's capabilities. The framework's core contribution lies primarily in orchestrating existing LLM capabilities through multi-agent collaboration rather than advancing the fundamental mechanisms of table reasoning.

For open-ended tasks like TAT-QA and WikiSQL, rigid evaluation metrics such as Exact Match and F1 scores may inadequately capture the flexible and varied responses that LLMs can generate. While these metrics provide standardized comparison points, they fail to acknowledge semantically equivalent answers that differ syntactically from ground truth. This challenge particularly affects our methodology, as scientific deliberation may produce valid alternative reasoning paths that yield correct but differently phrased answers.

Our evaluation lacks direct comparisons with other MAS for two primary reasons. First, our primary objective was to demonstrate the effectiveness of structured scientific methodology and multi-agent deliberation processes rather than achieving state-of-the-art performance across all benchmarks. Second, existing multi-agent approaches are often designed for specific reasoning scenarios with custom architectures that make fair, direct comparisons difficult \cite{tran_multi-agent_2025}. The lack of standardized, comparable benchmarks in the MAS field remains an open research challenge that the community has yet to address. We look forward to contributing to addressing this challenge.

Existing MAS exhibit several limitations that impact their effectiveness. These include inadequate task planning and allocation, instability during collaboration, insufficient self-validation mechanisms, and lack of error correction protocols \cite{cemri_why_2025, zhang_if_2025}. In our experiments, these challenges are to some extent observed. We acknowledge, however, that PanelTR has not fully overcome them. Future work could address these limitations by developing more flexible evaluation metrics that better capture semantic equivalence in responses, creating standardized benchmarks specifically designed to evaluate structured multi-agent reasoning processes, and exploring hybrid approaches that combine our scientific panel methodology with specialized parametric components for tasks requiring domain-specific expertise. Additionally, extending PanelTR to multimodal reasoning tasks involving tables, text, and images represents a promising direction for further research.

\section{Conclusion}
 
We introduced PanelTR, a multi-agent, scientist-persona discussion framework for complex table reasoning with individual \textit{Investigation, Self-Review}, and collaborative \textit{Peer-Review}, demonstrating how scientific methodology can transform complex table reasoning capabilities, and achieving remarkable reasoning capabilities without relying on task-specific training data. Experimental results across diverse benchmarks confirm that structured scientific inquiry produces substantial improvements over conventional approaches, particularly for tasks requiring nuanced analytical thinking.

The success of PanelTR highlights the potential of methodological frameworks in enhancing foundation model performance. Rather than pursuing increasingly larger models or extensive training datasets, our findings suggest significant gains can be achieved through careful orchestration of existing capabilities. This perspective presents a promising alternative pathway for advancing artificial intelligence systems facing complex reasoning challenges.
\section*{Acknowledgment}

This work is my first published paper and has been conducted independently. I am deeply grateful to the anonymous reviewers for their invaluable feedback and constructive suggestions that have significantly improved this work. Their guidance has been especially meaningful as this represents my first academic publication. I would also like to express my sincere gratitude to my father, Mr. Jianchao Ma, for his unconditional support throughout this journey.

\bibliographystyle{IEEEtran}
\bibliography{main}

\end{document}